# A Focal Any-Angle Path-finding Algorithm Based on A* on Visibility Graphs

Pei Cao, Zhaoyan Fan, Robert X. Gao, and J. Tang

*Abstract* **In this research, we investigate the subject of path-finding. A pruned version of visibility graph based on Candidate Vertices is formulated, followed by a new visibility check technique. Such combination enables us to quickly identify the useful vertices and thus find the optimal path more efficiently. The algorithm proposed is demonstrated on various path-finding cases. The performance of the new technique on visibility graphs is compared to the traditional A* on Grids, Theta* and A* on Visibility Graphs in terms of path length, number of nodes evaluated, as well as computational time. The key algorithmic contribution is that the new approach combines the merits of grid-based method and visibility graph-based method and thus yields better overall performance.**

## I. INTRODUCTION

The problem of finding the shortest path is frequently encountered in video games, robotics, GPS navigation, and path planning etc [1]~[7]. A recent research studied the method to obtain the shortest spatial path to plumbing multiple hydraulic components in additive manufacturing [8] where the plumbing length determines the weight, volume, and cost of product.

Known terrain pathfinding can be generally divided into two steps: 1) discretizing the continuous routing space, and 2) searching along the graph to find the path minimizing the cost value, which represents the overall length of the path from the starting node to the target node for a non-weighted graph. In this research, without loss of generality, square grids are used to discretize the terrain owing to its simplicity and popularity in various applications [9]. A review of different terrain discretizations can be found in [10].

Pei Cao is with the Mechanical Engineering Department, University of Connecticut, Storrs, CT, 06269 USA (e-mail: pei.cao@uconn.edu).

Zhaoyan Fan is with the Mechanical, Industrial and Manufacturing Engineering Department, Oregon State University, Corvallis, OR 97330 USA (e-mail: zhaoyan.fan@oregonstate.edu).

Robert X. Gao is with the Department of Mechanical and Aerospace Engineering, Case Western Reserve University, Cleveland, OH 44106 USA (e-mail: robert.gao@case.edu).

Jiong Tang is with the Mechanical Engineering Department, University of Connecticut, Storrs, CT, 06269 USA (e-mail: jtang@engr.uconn.edu).

A number of studies have been conducted on this subject. One of the earliest investigations was the Dijkstra's algorithm [11] where the cost value for the incremental search to the nearest goal is used. The well-known A* algorithm [12] made some improvements to Dijkstra's algorithm by adding heuristic cost from the current node to the target node to the evaluation. Due to its simplicity and guaranteed optimality guarantees, A* is widely used for solving path-finding problems as it is guaranteed mathematically to find a solution. However, the shortest path in gridded graph is not equivalent to the shortest path in continuous space where polynomials along the edge of grids can be replaced by straight lines [13]. One example is shown in Fig. 1. Fig. 1(a) illustrates the shortest graph path obtained by A* algorithm. Since A* algorithm can only take horizontal, vertical or diagonal propagation steps (i.e., zigzag steps) towards neighbor nodes, the path heading (direction) is constrained by propagation headings. The shortest continuous path is depicted in Fig. 1(b), where the path length is apparently shorter than that in Fig. 1(a). A commonly adopted solution to such a problem is to apply post-smoothing techniques to shorten the shortest grid path at the cost of an increase of computational time [14], [15]. The post processing techniques usually find a shorter any-angle path compared to A*, but it is not guaranteed to find the true shortest path [16]. Since they directly make adjustment to paths obtained by A* without knowing whether the post-processed path is optimal or not. These post processing techniques are often ineffective [17] due to the unavailability of the knowledge of the post processed paths.

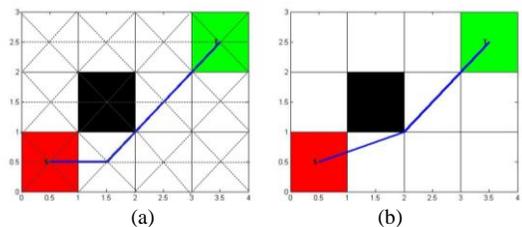

Fig. 1. The shortest graph path (a) vs. shortest continuous path (b)

Aiming at overcoming the limitations of A*, Field D* [17], an advanced version of D* Lite [18], was proposed which uses the linear interpolation of path costs of vertices to obtain any-angle paths. However, Field D* could result in paths that have unnecessary heading changes and should thus be smoothed further [19]. Some other approaches, such as Theta* [16], [20] and Lazy Theta* [21], were also proposed which embeds smoothing process into the A* searching to release the constraint on path searching directions (45 or 90 degrees).

Theta* finds shorter paths in less time than Filed D* according to literature [20]. Nevertheless, optimality is still not guaranteed [16], [22]. Fig. 2 shows the optimal path found by Theta* and the true shortest path. Similar to Theta*, Accelerated A* [23] is an any-angle pathfinding algorithm that is conjectured to be optimal but without strong theoretical argument. It can include a large proportion of nodes on the Closed List, i.e., large number of node evaluations for challenging problems [24].

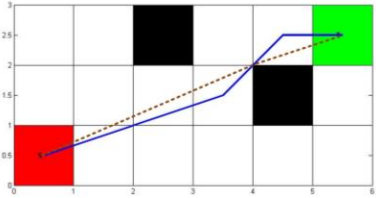

Fig. 2. Path found by Theta* vs. true shortest path

Although in application areas such as video games and navigations, the property of having shortest path length was not well emphasized compared to computational efficiency, it may become a critical issue in manufacturing scenarios. For example, the plumbing paths between components determine the total volume of product being manufactured, and accordingly, the weight, fabrication time, and cost. A common approach to finding the shortest paths in known terrain with polygonal obstacles is A* on Visibility Graphs [25]. But A* on Visibility Graphs suffers from inefficiency because it propagates along the edges of a visibility graph, the number of which grows exponentially in the size of grid. Even though the process can be accelerated by performing visibility check dynamically or using reduced visibility graphs [26], [27], A* on Visibility Graphs is still considered to be slow. Path-planning algorithms such as continuous Dijkstra and its variants [28], [29] as well as the recent Anya [24] also find shortest paths but have not yet been thoroughly evaluated experimentally. For more pathfinding methods, interested readers may refer to [10]. Generally speaking, to balance between distance and speed, an any-angle pathfinding algorithm should be designed to find a path shorter than that of A* meanwhile faster than A* on Visibility Graphs.

In this paper, a focal pruned visibility graph based on Candidate Vertices (*CV*) is developed, followed by a ray-casting based visibility check technique. The new approach reduces the number of evaluations needed for path-finding compared to both gird-based and visibility graph-based methods without sacrificing the optimality. The data structure associated is also introduced which helps to mitigate inefficiencies that come with A* on Visibility Graphs. The performance of the new algorithm, hereafter referred to as Focal Any-Angle A* (FA-A*) in this paper, is compared to A*, Theta* and A* on Visibility Graphs in terms of path length, nodes evaluated as well as computational time.

## II. CANDIDATE VERTICES (*CV*)

The Visibility Graph of a gridded map contains the starting node, the target node and all the vertices of obstacle blocks [25] (Lozano-Perez and Wesley, 1979). The reason why we need such a visibility graph is because the true shortest paths have heading changes only at the vertices of the blocks if any. But the vertices that the true shortest path may pass by are comprised of only a small subset of all vertices (Fig. 3). Those are the vertices that we really want to keep track of. Here we introduce a method to identify the possible vertices that the optimal path may pass through. We call this subset of vertices the Candidate Vertices (*CV*). All the Candidate Vertices along with starting node and target node and their edges constitute the focal pruned visibility graph. The method is explained in detail below which includes a one-time preprocessing. The *CV* set is constructed dynamically in every step during the search.

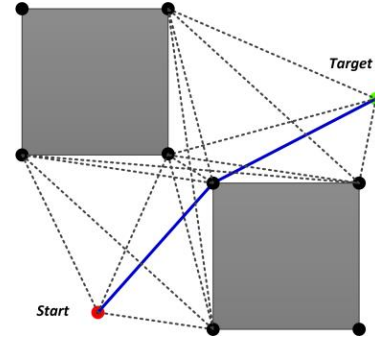

Fig. 3. Visibility Graph of two nodes and two obstacles, and the shortest path

### A. Preprocessing

For a problem with *n* obstacles, an *n* by *n* symmetric proximity matrix ***D*** is used to record pairwise distances. For example, the proximity value between the *i*-th obstacle and the *j*-th obstacle is specified by the distance between them if obstacles are unit squares,

$$\boldsymbol{D}(i, j) = |i, j| \qquad (1)$$

Then the obstacles that have proximity values smaller than $\sqrt{2}$ will be clustered if diagonal move between obstacles is allowed. If not, we cluster the pairs with proximity values less of equal to $\sqrt{2}$ (Fig. 4). Here single linkage method is applied for clustering obstacle points [31].

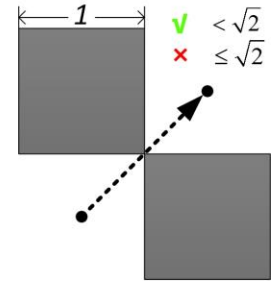

Fig. 4. Diagonal move in between obstacles

Next, for each cluster, quick hull algorithm (Barber et al., 1995) is adopted to obtain the vertices on the convex hull of each cluster. It is possible that when non-convex clusters are tightly placed against each other like mortise and tenon, the optimal path may go through the vertices that are not on the convex hull of the clusters. Thus, it is recommended to consider all the vertices on the convex corners of a cluster when the proportion of obstacles is large or when we are dealing with maze routing problems. However, for obstacle-avoidance problems, vertices on convex hull are adequate and thus will be adopted in the first three numerical tests in this paper (Tests A, B and C in Section V). In the fourth numerical test (Test D), vertices on the convex corners will be considered. After eliminating the vertices on boundaries, the remaining vertices are the vertices that will be further used to construct the *CV*, and we call this group of vertices $V_1$. An example pathfinding problem after preprocessing can be found in Fig. 5(a) where 36 vertices are in the pool for Candidate Vertices. The preprocessing only needs to be performed once for each gridded map regardless the positions of starting node and target node. On the other hand, full visibility graphs need to be constructed for different starting node and target node setups.

One merit of this preprocessing technique is that, no matter how small the grid size is or how many obstacles are used to capture the silhouette of one blocking feature, it can be always regarded as one cluster instead of a large number of obstacles. It enhances the computational performance of the algorithm along the line when maps have more and more details.

B. *Generation of Candidate Vertices (CV)*

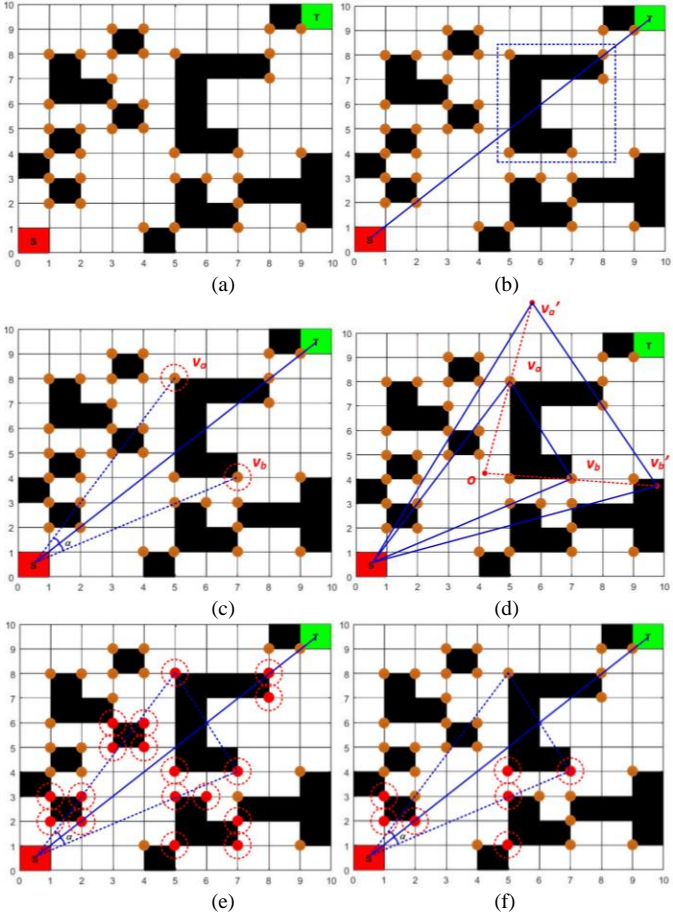

(a) Vertices in $V_1$ after (b) Blocking cluster and $V_2$ (inside the box)
(c) Vertices corresponding to the largest acute angle (d) Triangle enlarging
(e) Vertices in $V_3$ (f) Candidate Vertices (marked with dotted circles)
Fig. 5. An example of finding Candidate Vertices

After preprocessing, the Candidate Vertices (*CV*) will be dynamically generated throughout the searching process. In each step, the clusters that block the straight line between the current node and the target node are considered first. In other words, we first investigate the vertices on convex hulls of the clusters that block the straight line between the current node and the target node.

Fig. 5 illustrates the steps of finding Candidate Vertices in FA-A*. As shown in Fig. 5(b), the dotted box indicates the blocking cluster and associated vertices. We name this set of vertices $V_2$ which is a subset of $V_1$. The visibility is checked to determine whether any cluster blocks the straight line between two nodes. The visibility check technique adopted will be introduced in the following section. Next, we find the vertex on each side of the straight line from $V_2$ that constructs the largest acute angle ($\alpha$) with the current node and the target node, as shown in Fig. 5(c). The vertex on each side that corresponds to the largest acute angel is represented by $v_a$ and $v_b$ respectively. It is acute because if a blocking cluster stretches from one end to the other, our focus to the target will be blurred. Implicitly, clusters that stay close to the current node will have higher priority in terms of containing $v_a$ and $v_b$. Assume the current node is represented by $p$, the target node is $q$ and the vertex is $v$. We use the linear equation of straight line to determine which side $v$ falls into. The straight line defined by $p$ and $q$ is,

$$Ax + By + C = 0 \qquad (2)$$

where $A = p_y - q_y$, $B = q_x - p_x$, and $C = p_x \cdot q_y - q_x \cdot p_y$.

Substituting the position of the vertex into the above equation, we have,

$$D = Av_x + Bv_y + C \qquad (3)$$

The position of the vertex $v$ with respect to the straight line can be determined by comparing $D$ to 0. Then we can calculate the angel $\alpha$ respectively for vertices on each side, i.e.,

$$\alpha = \arccos \frac{\overrightarrow{pq} \cdot \overrightarrow{pv}}{|\overrightarrow{pq}||\overrightarrow{pv}|} \qquad (4)$$

We now have a triangle that is defined by the current node, $v_a$, and $v_b$. An extreme case is when $v_a$, $v_b$ are collinear and, consequently, no triangle or polygon can be formed. In that case only the vertices in $V_1$ that are on the line segment between $v_a$ and $v_b$ will be considered. Additionally, if all the vertices in $V_2$ are at the same side of the straight line, $v_a$ and $v_b$ then correspond to the largest and smallest acute angles, respectively. We also introduce a technique that enlarges the triangle by pushing $v_a$, and $v_b$ away from the barycenter $o$ of the triangle. The updated $v_a$, and $v_b$ can be obtained as

$$v_a' = o + w \cdot \overrightarrow{ov_a} \qquad (5)$$

$$v_b' = o + w \cdot \overrightarrow{ov_b} \qquad (6)$$

where $w$ is a scale factor. The default value of $w$ is 1 which means the triangle stays intact. Fig. 5(d) shows the comparison between $v_a$, $v_b$ and $v_a'$, $v_b'$ when $w$ is 2.

The vertices in $V_1$ that belong to the clusters inside or tangential to this triangle are narrowed down as the vertices set $V_3$ (Fig. 5(e)). Testing whether a point is inside a polygon is a straightforward operation. In this research, we use the algorithm proposed in [32]. Finally, we proceed to check if the vertices in $V_3$ are visible from the current node; all the visible ones are the so-called Candidate Vertices (*CV*). The *CV*, along with the current node, the target node and their edges constitute the focal pruned version of visibility graphs (Fig. 5(f)). Then the *CV* are used in A* algorithm as neighboring nodes instead of the nodes adjacent to the current node.

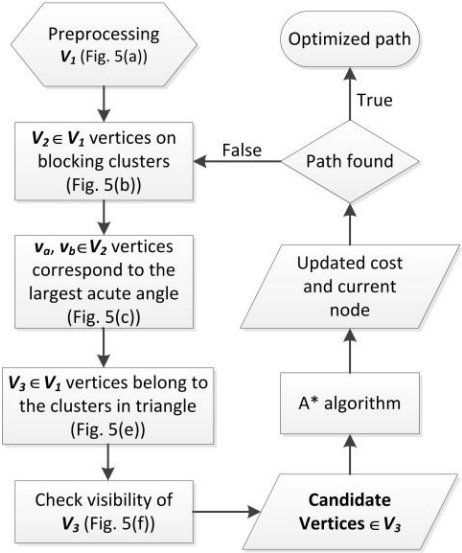

Fig. 6. Flowchart of finding the Candidate Vertices as part of pathfinding

A flowchart is drawn to articulate the process of determining the Candidate Vertices iteratively as a part of pathfinding algorithm in Fig. 6. With the procedures mentioned above, we are able to reduce the computational burden of A* on Visibility Graphs through checking the visibility of vertices in $V_3$ instead of all the vertices and only propagating to *CV* whereas the key idea of A* on Visibility Graphs has been preserved.

## III. VISIBILITY CHECK

In Theta*, a visibility check technique called line-of-sight is adopted which is derived from [33]. The idea of line-of-sight check is straightforward. It checks certain grids between two nodes based on their relative position. If the examined grid is an obstacle, the two nodes are non-visible to each other. Line-of-sight checks can be performed efficiently with only integer operations on square grids. However, such technique is ad-hoc when adapted to other discretizations or different searching rules because the grids to examine are different case by case. Thus, in this research, we propose a ray-casting algorithm to check for visibility between two nodes based on the algorithm proposed in [30]. The computational complexity of the two approaches is similar ($O(n)$), but the approach used in this paper is more flexible and can accommodate various applications systematically.

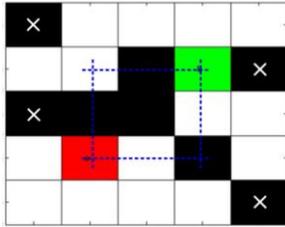

Fig. 7. The obstacles in between starting node and target node

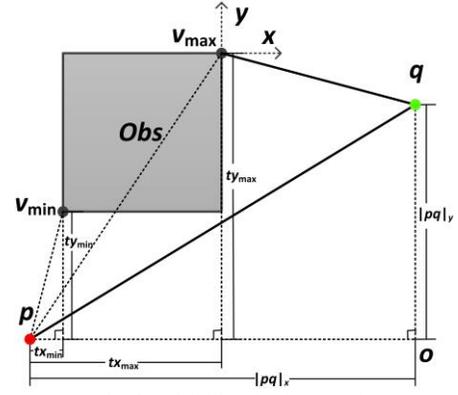

Fig. 8. Visibility check example

The steps of checking the visibility between two nodes with the existence of obstacles are outlined below:

*1) Consider the obstacles occurring between the nodes*

Observe Fig. 7. When an obstacle occurs between the nodes, the *x*, *y* values of the obstacle in Cartesians system cannot exceed the maximum or minimum *x*, *y* values of the two nodes. The obstacles marked with a cross in Fig. 7 should be eliminated. If two nodes have the same *x* value or *y* value, then the visibility can be determined directly by examining the grids in between.

*2) Check if the obstacles block the ray between nodes*

We check the obstacles that are bounded by the nodes (Fig. 7) one after another to see if any of them intersect with the line segment between the two nodes. Consider Fig. 8 as an example. *p* and *q* are two nodes, and *Obs* is a unit obstacle where the vertices with the smallest and largest *x*, *y* values are $v_{min}$ and $v_{max}$, respectively. If

$$\frac{tx_{\min}}{\left|(\overrightarrow{pq})_x\right|} < \frac{ty_{\max}}{\left|(\overrightarrow{pq})_y\right|} \tag{6}$$

and

$$\frac{ty_{\min}}{\left|(\overrightarrow{pq})_y\right|} < \frac{tx_{\max}}{\left|(\overrightarrow{pq})_x\right|} \tag{7}$$

then the line segment connecting *p* and *q* goes through the obstacle (intersect). In Eqs. (6) and (7), $\left|(\overrightarrow{pq})_x\right|$, $\left|(\overrightarrow{pq})_y\right|$ are projections of |*pq*| on *x*-axis and *y*-axis, $tx_{min}$, $ty_{min}$ are projections of |$pv_{min}$| on *x*-axis and *y*-axis, and $tx_{max}$, $ty_{max}$ are projections of |$pv_{max}$| on *x*-axis and *y*-axis, as shown in Fig. 8. If we change '<' to '≤' in Eqs. (6) and (7), merely passing through a vertex will be considered as intersection.

Once an obstacle is calculated to have intersected with the line segment between the nodes, we can terminate the examination and conclude that the two nodes are non-visible to each other. For more details, interested readers may refer to [30].

## IV. FOCAL ANY-ANGLE A* ON VISIBILITY GRAPHS

### A. FA-A* Algorithm

We propose the approach to reduce the visibility graph and check visibility. The A* algorithm that uses such techniques is referred to as FA-A* (Focal Any-Angle A*). The pseudo code for FA-A* is provided below:

```
Main()
  open = ∅
  g(n_start) := 0
  parent(n_start) = n_start
  open.Insert(n_start, parent(n_start), g(n_start), h(n_start))
  n_current = n_start
  While n_current exists
    propagate = CV of n_current
    if n_target ∈ propagate
      open.Insert(n_target, n_curren, g(n_curren)+c(n_curren, n_target),0)
      return "path found"
    end if
    open.Refresh (propagate, n_current)
    open.Close(n_current)
    n_current = the node in open with the smallest g+h
  End while
end

open.Refresh(propagate, n_current)
  for every n' ∈ propagate
    g(n')_new = g(n_current)+c(n_current, n')
    if n' ∈ open
      if g(n')_new < g(n') value in open
        open.Update(n', n_current, g(n')_new, h(n'))
      end if
    else
      open.Insert(n', n_current, g(n')_new, h(n'))
    end if
  end for
end
```

In the pseudo-code, $n$ means node. Therefore $n_{start}$, $n_{target}$, $n_{current}$, are the starting node, the target node, and current node respectively. The notation $g(n)$ represents the shortest path length from the starting node to node $n$ found so far. $c(n_1, n_2)$ is the cost travel from $n_1$ to $n_2$, i.e., the Euclidean distance between $n_1$ and $n_2$. And $h(n)$ is the heuristic cost from node $n$ to target node. In this research, without loss of generality, the heuristic cost is approximated by the Euclidean distance between two nodes.

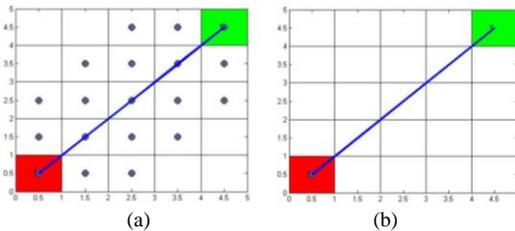

Fig. 9. Node evaluations comparison between A* on grids and FA-A*

The key difference between FA-A* and A* is that FA-A* only propagates to Candidate Vertices (**CV**) of the current node instead of all the neighboring nodes (please see the grayed line in the pseudo-code). Consider a simple case where there is no obstacle. A* or Theta* need to evaluate 17 nodes along the search until they find the target node (Fig. 9(a)), whereas FA-A* could find the target node directly from starting node because the target node is one of the Candidate Vertices of the starting node (Fig. 9(b)). The evaluated nodes are marked with dots in Fig. 9. Note that in the pseudo-code, we only maintain an *open* list, whereas in A* and Theta* a *closed* list co-exists with the *open* list. Such change is facilitated by using the *open.Close* operation which prevents the present current node from being either the current node or appearing in the **CV** again in the future. This change may help to enhance the performance of FA-A* in terms of efficiency.

### B. Optimality

The selection process of Candidate Vertices proposed could be regarded as a greedy approach to reduce the number of visibility checks and function evaluations compared to the methods that use complete visibility graphs.

The algorithm is inspired by the scenarios when two nodes are connected with rubber band as the optimal rubber band connections always have the shortest distance possible and have heading changes mostly at or in between the vertices that belong to the clusters that play the biggest roles in blocking the two nodes. The role played by the obstacles in blocking two nodes is an abstract notion which is quantified in our research as $\alpha$. Then $v_a$ and $v_b$ correspond to the largest outreaches of the blocking clusters on each side. By specifying $V_3$, we aim to pinpoint all the vertices that the optimal path may pass through. While the algorithm finds better paths than A* and Theta*, the optimality is not always guaranteed for some highly discrete cases. These will be illustrated in numerical Tests A and B. Nevertheless, the proposed method finds paths that are very close to the shortest paths. That is why we introduce a scale factor $w$ to enlarge the triangle, i.e., to include more vertices in $V_3$ at the cost of worse computational efficiency. To begin with, we try to reduce the vertices considered compared to complete visibility graphs. Increasing the value of $w$ will gradually recede such reduction effect and enable the algorithm to find the shortest path in particular situations. As the numerical test results in Section V indicates, in most cases, a default value 1 for $w$ is good enough for the algorithm to find the shortest path.

### C. Computational Complexity Analysis

This section provides the complexity analysis of FA-A*. The operations of FA-A* and their worst-case complexity are outlined as follows.

1) Prepossessing: $O(|\mathbf{Obs}| \log |\mathbf{Obs}|)$ as it is the complexity for quick hull algorithm. $|\mathbf{Obs}|$ is the number of obstacles.

2) Check which clusters block the straight line between starting node and current node: $O(n)$. $O(n)$ is the complexity for visibility check where $n$ represents the number of obstacles in between the two nodes.

3) Calculate $\alpha$: $O(|V_2|)$. $|V_2|$ is the size of $V_2$.

4) Check if a vertex is inside the triangle: $O(3|V_1|)$. $O(3)$ is the complexity to check if a point is inside of a triangle.

5) Check if vertices in $V_3$ are visible from current node: $O(n|V_3|)$.

In summary, the computational complexity for FA-A* is

$$O(\text{FA-A*}) = O(|Obs|\log |Obs|) \\ + O(Expan_{\text{FA-A*}}) \times O(n+|V_2|+3|V_1|+n|V_3|) \quad (8)$$

where $Expan_{\text{FA-A*}}$ is the number of expansion in FA-A*. This can be simplified to

$$O(\text{FA-A*}) = O(Expan_{\text{FA-A*}} \times (n+|V_2|+3|V_1|+n|V_3|)) \quad (9)$$

The first term in Eq. 8 is omitted because that the second term has the dominant effect on complexity as the scale of the problem increases.

Meanwhile, the computational complexity of A* is $O(8 \times Expan_{A^*})$, the complexity of Theta* is $O((8+n) \times Expan_{\theta^*})$, and the complexity of A* on Visibility Graphs is $O(n|V| \times Expan_{A^* \text{on} V})$ where $|V|$ is the number of vertices. The number of expansion for each algorithm has the following empirical relations,

$$Expan_{A^*} > Expan_{\text{FA-A*}} \quad (10)$$

$$Expan_{\theta^*} > Expan_{\text{FA-A*}} \quad (11)$$

$$Expan_{A^* \text{on} V} > Expan_{\text{FA-A*}} \quad (12)$$

Since $V$ is a set of all the vertices, we have,

$$n|V| > n+|V_2|+3|V_1|+n|V_3| \quad (13)$$

FA-A* can be regarded as a light version of A* on Visibility Graphs in terms of computational complexity, given Eqs. (12) and (13). Since the sizes of $V_1$, $V_2$ and $V_3$ are proportional to the number of clusters, FA-A* is efficient in solving problems with considerable clusters of obstacles rather than discrete small obstacles with the same proportion of obstacles. This will be validated in numerical Test C in Section V. As a result, FA-A* has advantages over methods that propagate to immediate neighbors like A* and Theta* computationally when handling maps that have small number of clusters of obstacles. Meanwhile, FA-A* is always more efficient than A* on Visibility Graphs. All of the above could make FA-A* useful in solving either obstacle avoidance or maze routing problems.

### D. Data Structure

In this research, we maintain a concise data structure where two main matrices are used. The first matrix, denoted as $V_{\text{all}}$, is a $|V|$ by 4 matrix, where $|V|$ is the number of all vertices. Each row of $V_{\text{all}}$ corresponds to one vertex in $V$ where the first two columns are the position of the vertex in the coordinate system, the third column is the obstacle index to which the vertex belongs, and the last column is the cluster index to which the obstacle belongs. Similarly, the second matrix, denoted as $V_{\text{convex}}$, is a $|V_1|$ by 4 matrix where each row corresponds to one vertex in $V_1$.

$V_{\text{all}}$ is used to determine $V_1$ in preprocessing, and $V_{\text{convex}}$ is formed accordingly. After using $V_{\text{convex}}$ to determine $V_2$, we use $V_{\text{all}}$ to check whether a cluster is inside or tangential to the triangle. Finally, $V_{\text{convex}}$ can be used to determine $V_3$.

## V. NUMERICAL TESTS

In this section, we evaluate the performances of four algorithms, i.e., A* on Grids (A* on G), Theta*, A* on Visibility Graphs (A* on V), and FA-A* under four representative test cases. In Tests A and B, we vary the location of the starting node and the target node, as well as the complexity of the map. A total number of 15 maps are used to assess the performance of each algorithm with respect to 2D benchmark cases. Next in Test C, we maintain the number of obstacles while changing the number of clusters to investigate the effect of the number of clusters to path-finding outcomes. Finally, in Test D we examine the algorithms on a maze-routing problem.

### A. Placing starting and target nodes in the corners

In this sub-section, 10 maps with different complexities are used when the starting node and the target node are placed respectively at the bottom left corner and the top right corner. The A* on Visibility Graphs used is optimized to the extent that we only consider the vertices in the convex corners when checking visibility. The metrics employed are the length of the path ($L$), the number of node evaluations (*No.*), and the computational time ($T$). The scale factor $w$ for FA-A* is set to the default value 1 unless FA-A* cannot find the shortest path. We then increase $w$ gradually with a step size of 0.1 until the shortest path is found. In this case, the smallest $w$ for the true shortest path to be found and the corresponding performance are recorded and compared as well. The $h$-value used for the experiments are the Euclidean distance from the current node to the target node. The computational time is the average over 5 runs. All algorithms are implemented in MATLAB and executed on a 2.40 GHz 2 processors (Xeon E5620) desktop. The runtime could be improved by using other programming languages.

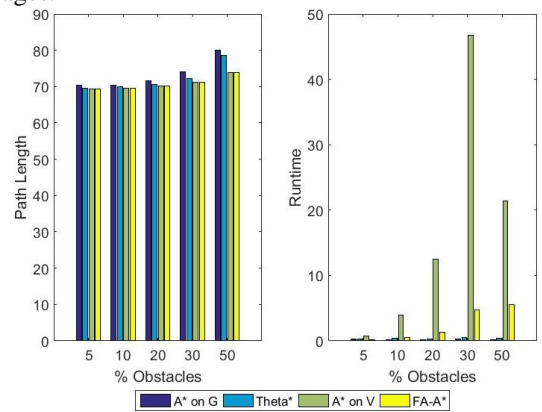

Fig. 10. Random 50*50 maps with different proportion of obstacles

We start off by comparing the algorithms on 50*50 gridded graphs. The obstacles are randomly generated with 5%, 10%, 20%, 30% and 50% fraction of the map respectively. The test results are plotted in Fig. 10 and listed in Table 1. FA-A* always finds the shortest path except that the proportion of obstacles is 10% where the shortest path found is 0.543% longer than true shortest path. As demonstrated, FA-A* outperforms the A* and Theta* in all obstacle scenarios in terms of path length and number of node evaluations. When the number of obstacle increases, FA-A* slows down because of *CV* calculations. It is always faster than A* on Visibility Graphs, and faster than A* on Grids and Theta* when the proportion of obstacles is 5%. Fig. 11 shows the path/shortest path versus the runtime which demonstrates how the algorithms balance between shortest distance and shortest runtime. Figs. 11(a), (c), (e) and (i) show that FA-A* stays in the left bottom corner indicating a good compromise of the grid-based technique and the visibility graph-based technique. For the case with 5% obstacles, as can be seen in Fig. 11(a), FA-A* prevails in all metrics.

Table 1 Performance on 50*50 graphs

| Map Info | Algorithm | L | No. | T |
|---|---|---|---|---|
| 50*50 5% | A* on G | 70.4680 | 487 | 0.2964 |
| | Theta* | 69.5051 | 406 | 0.3120 |
| | A* on V true shortest | 69.4394 | 392 | 0.7956 |
| | FA-A*($w$=1) | 69.4394 | 31 | 0.1404 |
| 50*50 10% | A* on G | 70.4680 | 416 | 0.1716 |
| | Theta* | 69.9440 | 505 | 0.4056 |
| | A* on V true shortest | 69.5763 | 705 | 3.9156 |
| | FA-A*($w$=1) | 69.6141 | 114 | 0.4836 |
| | FA-A*($w$=1.6) | 69.5763 | 122 | 0.4992 |
| 50*50 20% | A* on G | 71.6396 | 483 | 0.2028 |
| | Theta* | 70.6594 | 429 | 0.3120 |
| | A* on V true shortest | 70.2469 | 828 | 12.4997 |
| | FA-A*($w$=1) | 70.2469 | 207 | 1.3232 |
| 50*50 30% | A* on G | 73.9828 | 668 | 0.2808 |
| | Theta* | 72.1964 | 551 | 0.4680 |
| | A* on V true shortest | 71.2623 | 839 | 46.7961 |
| | FA-A*($w$=1) | 71.2623 | 353 | 4.7736 |
| 50*50 50% | A* on G | 79.9828 | 589 | 0.2184 |
| | Theta* | 78.5568 | 519 | 0.4056 |
| | A* on V true shortest | 73.9302 | 358 | 21.3549 |
| | FA-A*($w$=1) | 73.9302 | 240 | 5.5632 |

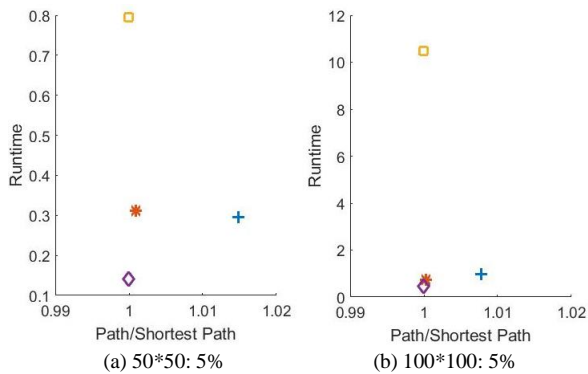

(a) 50*50: 5%  (b) 100*100: 5%

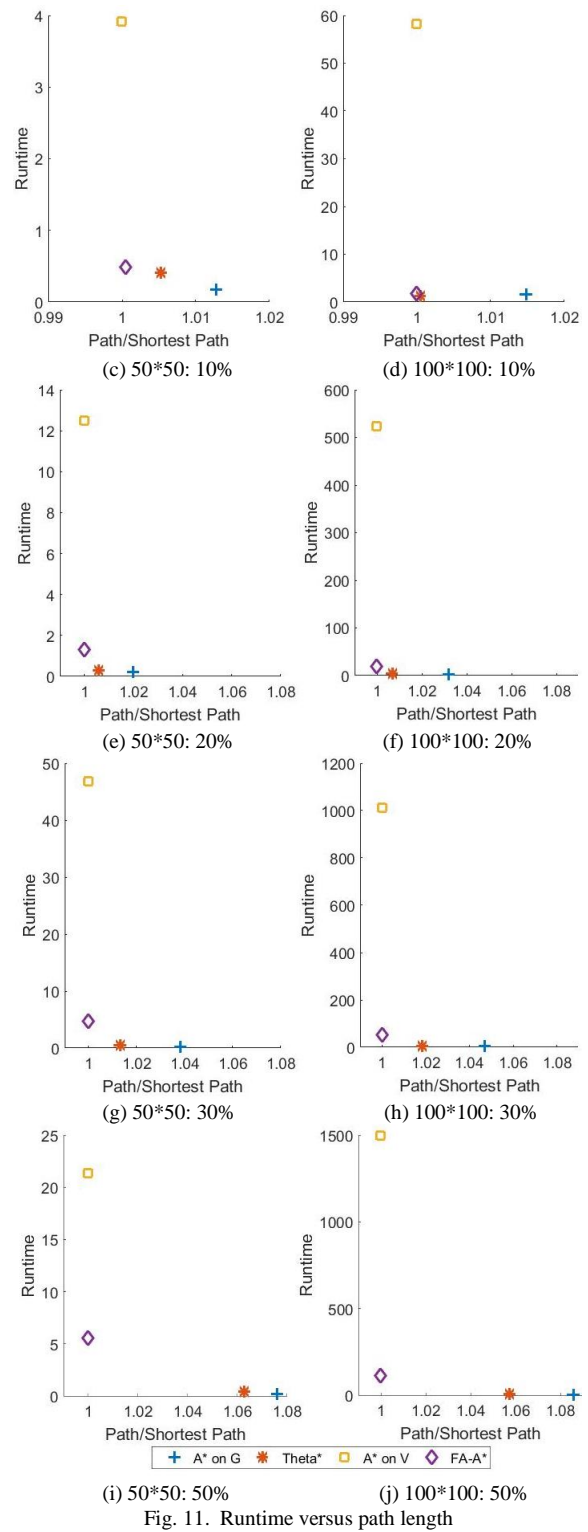

(c) 50*50: 10%  (d) 100*100: 10%
(e) 50*50: 20%  (f) 100*100: 20%
(g) 50*50: 30%  (h) 100*100: 30%
(i) 50*50: 50%  (j) 100*100: 50%

Fig. 11. Runtime versus path length

We then proceed to 100*100 graphs with the same obstacles fractions as in 50*50 cases. The results are reported in Fig. 2 and visualized in Fig. 12. A demonstration of paths found by each algorithm and the comparison can be found in Fig. 13 where the evaluated nodes are filled with dots. Our observation from this set of experiments is consistent with 50*50 cases. When there are a small number of obstacles, FA-A* not only finds the best path, but also excels in computational speed. When the obstacle fraction increases, the runtime sacrifices; in return, the paths found by FA-A* are shorter than those of A* and Theta* significantly. Such observations are also supported by Fig. 11. In the test case where obstacles occupy 30% of the map, FA-A* is unable to find the shortest path until $w$ is increased to 2.2. The path found by FA-A* is compared to the true shortest path in Fig. 14. The only difference is zoomed in while the rest of two paths coincide with each other. This happens because when $w=1$, FA-A* fails to include the vertex as Candidate Vertices in one particular step which is solved by increasing $w$ to 2.2 if optimality is highly emphasized. Nevertheless, FA-A* with $w=1$ still finds better path than A* or Theta* as it is only 0.007428% longer than the true shortest path. Usually such difference is acceptable. Based on our experiment results, the choice of 1 for $w$ should be adequate for the most of the time.

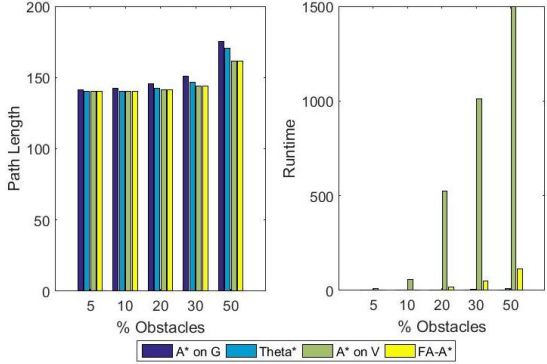

Fig. 12. Random 100*100 maps with different proportion of obstacles

Table 2 Performance on 100*100 graphs

| Map Info | Algorithm | L | No. | T |
|---|---|---|---|---|
| 100*100 5% | A* on G | 141.1787 | 950 | 0.9828 |
| | Theta* | 140.1210 | 723 | 0.7332 |
| | A*on V true shortest | 140.0883 | 1345 | 10.4793 |
| | FA-A*($w=1$) | 140.0883 | 62 | 0.4680 |
| 100*100 10% | A* on G | 142.3503 | 1364 | 1.5756 |
| | Theta* | 140.3394 | 958 | 1.2948 |
| | A*on V true shortest | 140.2633 | 2365 | 58.3068 |
| | FA-A*($w=1$) | 140.2633 | 193 | 1.7964 |
| 100*100 20% | A* on G | 145.8650 | 2344 | 3.1668 |
| | Theta* | 142.3683 | 1653 | 3.4632 |
| | A*on V true shortest | 141.3687 | 2625 | 524.1037 |
| | FA-A*($w=1$) | 141.3687 | 662 | 19.3401 |
| 100*100 30% | A* on G | 150.7939 | 2658 | 3.4788 |
| | Theta* | 146.7182 | 2029 | 5.4756 |
| | A*on V true shortest | 144.0496 | 2280 | 1011.5 |
| | FA-A*($w=1$) | 144.0603 | 1190 | 50.9444 |
| | FA-A*($w=2.2$) | 144.0496 | 1305 | 57.0032 |
| 100*100 50% | A* on G | 175.3797 | 2889 | 3.6284 |
| | Theta* | 170.8014 | 2706 | 9.1977 |
| | A*on V true shortest | 161.5234 | 2193 | 1498.0 |
| | FA-A*($w=1$) | 161.5234 | 1491 | 112.6314 |

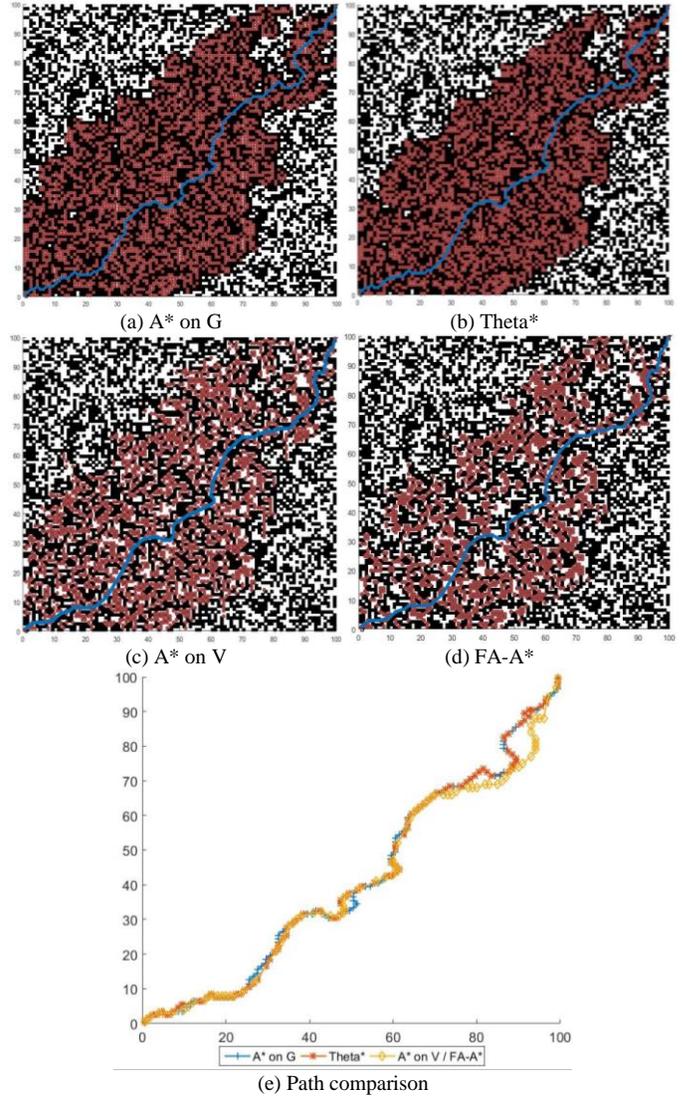

(a) A* on G  (b) Theta*

(c) A* on V  (d) FA-A*

(e) Path comparison

Fig. 13. Paths found and nodes evaluated by each algorithm (100*100, 50%)

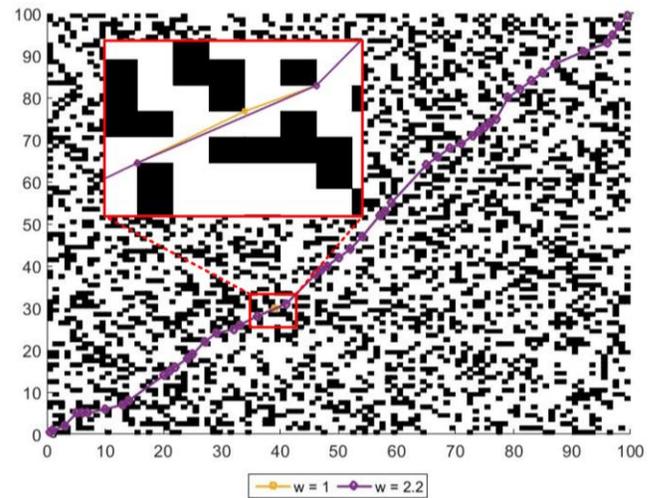

Fig. 14. Paths comparison of FA-A* with different $w$ (100*100, 50%)

B. *Placing starting node in middle*

Next we apply the algorithms to 300*300 maps with various proportions of obstacles. The starting node is placed in the middle of the map (150, 150), and the target node is randomly placed. The test results are visualized in Fig. 15 and recorded in Table 3. Because it took A* on Visibility Graphs too long to find the optimal paths for last two cases, the results are omitted. The performance comparison of the other three algorithms is our focus here. FA-A* always finds better paths than A* on Grids and Theta*. Moreover, FA-A* shows the best performance in all three metrics when the fraction of obstacles are 5% and 10%.

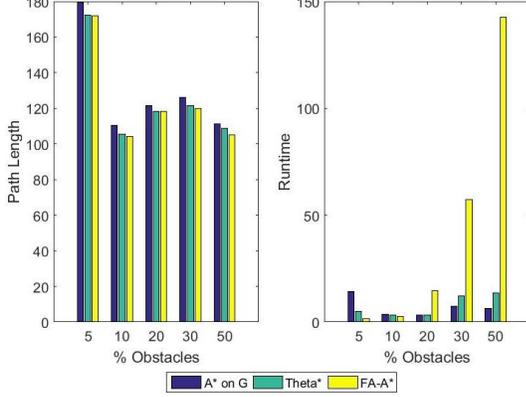
Fig. 15. Random 300*300 maps with different proportion of obstacles

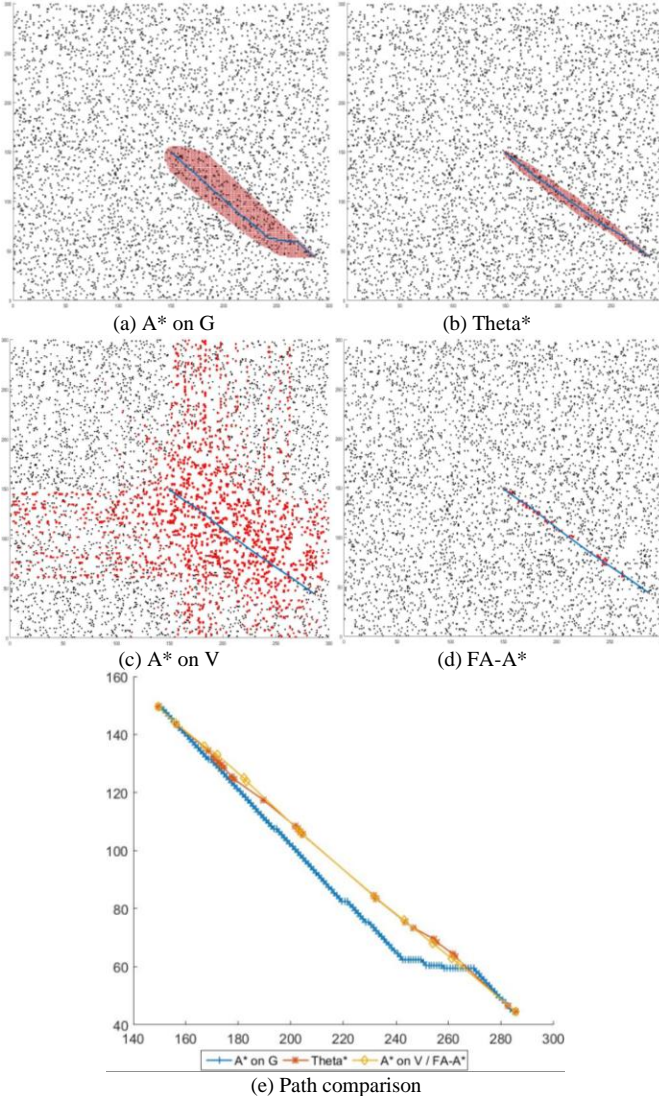
(a) A* on G  (b) Theta*
(c) A* on V  (d) FA-A*
(e) Path comparison
Fig. 16. Paths found and nodes evaluated by each algorithm (300*300, 10%)

Similar to A* on Visibility Graphs, the computational time of FA-A* increases as the number of obstacles increases. A comparison of paths is provided in Fig. 16, where A* on Grids evaluates a large number of nodes (4911) around the path found and Theta* managed to reduce the number of node evaluations on the basis of A* on Grids. Meanwhile, A* on Visibility Graphs evaluates 4589 nodes regardless the position of the target as shown in Fig. 16, and FA-A* only evaluates 137 nodes in the searching process which is significantly better than other three algorithms.

Table 3 Performance on 300*300 graphs

| Map Info | Algorithm | $L$ | $No.$ | $T$ |
|---|---|---|---|---|
| 300*300 5% Target (286, 45) | A* on G | 179.4924 | 4911 | 14.1165 |
| | Theta* | 172.3074 | 1619 | 4.9140 |
| | A*on V true shortest | 171.9521 | 4589 | 171.9521 |
| | FA-A*($w$=1) | 171.9521 | 137 | 1.7160 |
| 300*300 10% Target (170, 48) | A* on G | 110.2843 | 2090 | 3.7284 |
| | Theta* | 105.3346 | 967 | 3.1512 |
| | A*on V true shortest | 104.3420 | 4470 | 1826.2 |
| | FA-A*($w$=1) | 104.3543 | 154 | 2.5428 |
| | FA-A*($w$=1.6) | 104.3420 | 168 | 2.8704 |
| 300*300 20% Target (230, 236) | A* on G | 121.4802 | 1359 | 3.1824 |
| | Theta* | 118.4848 | 757 | 3.3384 |
| | A*on V true shortest | 118.1886 | 2991 | 8894.4 |
| | FA-A*($w$=1) | 118.1886 | 389 | 14.6573 |
| 300*300 30% Target (50, 212) | A* on G | 126.2670 | 2211 | 7.5192 |
| | Theta* | 121.5660 | 1391 | 12.2617 |
| | A*on V true shortest | N/A | N/A | N/A |
| | FA-A*($w$=1) | 119.9381 | 759 | 57.1432 |
| 300*300 50% Target (75, 111) | A* on G | 111.2965 | 1327 | 6.2712 |
| | Theta* | 108.6661 | 1217 | 13.6813 |
| | A*on V true shortest | N/A | N/A | N/A |
| | FA-A*($w$=1) | 105.0499 | 706 | 142.4913 |

The results from the three sets of experiments above are consistent in the sense that FA-A* outperforms three other algorithms in all categories when the number of obstacles is small. However, the computational time performance of FA-A* deteriorates as the number of obstacles increases. Based on our complexity analysis, the main factor that affects the computational time of FA-A* is the number of clusters rather than the number of obstacles. Naturally, when the obstacles are randomly generated, the number of clusters is proportional to the number of obstacles because they are more discretely placed.

*C. Different number of clusters*

Here we investigate the impact of the number of clusters. In many applications, the obstacles are clustered together to capture the details of the blocking features. Thus, in this test, we use 50*50 maps with 30% of obstacles but the number of clusters is confined. The position, size and shape of the cluster are all randomly generated. The test results are reported in Table 4 and Fig. 17. As can be observed, FA-A* always finds the shortest path and the time used decreases as the number of clusters increases. The runtime of A* on G and Theta* do not show much of an improvement. A comparison of the four algorithms with varying numbers of clusters is illustrated in Fig. 18. When the number of clusters is 35 (Fig. 19), FA-A* dominates in all categories. In an extreme case when there is only one obstacle, FA-A* finds the shortest path and takes 0.0268 longer than A* on G because of the overhead introduced by the calculation of Candidate Vertices.

Evidently, visibility graph based methods are superior to methods that propagate to immediate neighbors for problems with small number of clusters. Therefore, they are more practical towards a series of applications such as car obstacle avoidance and plumbing designs.

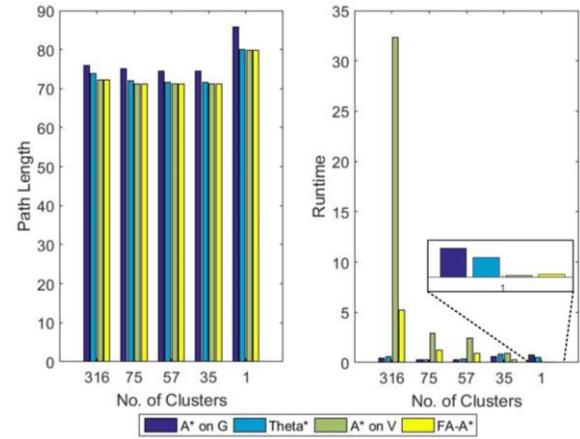

Fig. 17. Random 50*50 maps with different number of clusters

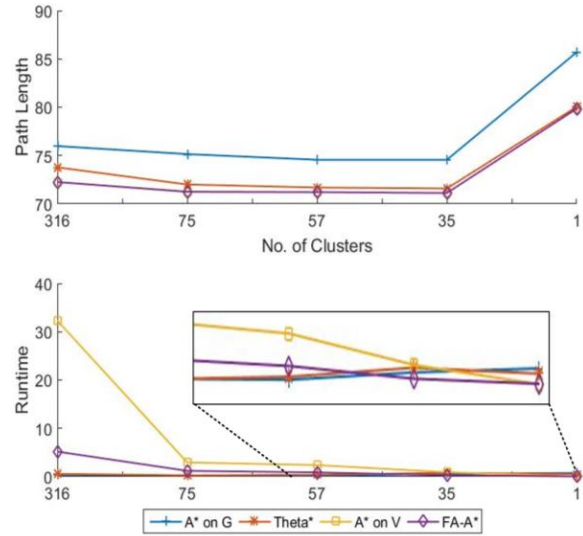

Fig. 18. Performance comparison (50*50)

Table 4 Performance on 50*50 graphs

| Map Info | Algorithm | $L$ | No. | $T$ |
|---|---|---|---|---|
| 50*50 30% 316 clusters | A* on G | 75.9828 | 793 | 0.4056 |
| | Theta* | 73.7824 | 633 | 0.5928 |
| | A* on V true shortest | 72.2600 | 741 | 32.3234 |
| | FA-A*($w$=1) | 72.2600 | 372 | 5.1948 |
| 50*50 30% 75 clusters | A* on G | 75.1543 | 687 | 0.2964 |
| | Theta* | 72.0079 | 401 | 0.2808 |
| | A* on V true shortest | 71.2361 | 254 | 2.9484 |
| | FA-A*($w$=1) | 71.2631 | 102 | 1.2168 |
| 50*50 30% 57 clusters | A* on G | 74.5685 | 594 | 0.2496 |
| | Theta* | 71.6930 | 453 | 0.3900 |
| | A* on V true shortest | 71.2190 | 178 | 2.4024 |
| | FA-A*($w$=1) | 71.2190 | 84 | 0.8892 |
| 50*50 30% 35 clusters | A* on G | 74.5685 | 946 | 0.5928 |
| | Theta* | 71.5985 | 743 | 0.8112 |
| | A* on V true shortest | 71.1266 | 124 | 0.9360 |
| | FA-A*($w$=1) | 71.1266 | 36 | 0.2964 |
| 50*50 30% 1 cluster | A* on G | 85.6985 | 1403 | 0.7800 |
| | Theta* | 80.0536 | 762 | 0.5304 |
| | A* on V true shortest | 79.8618 | 4 | 0.0356 |
| | FA-A*($w$=1) | 79.8618 | 4 | 0.0624 |

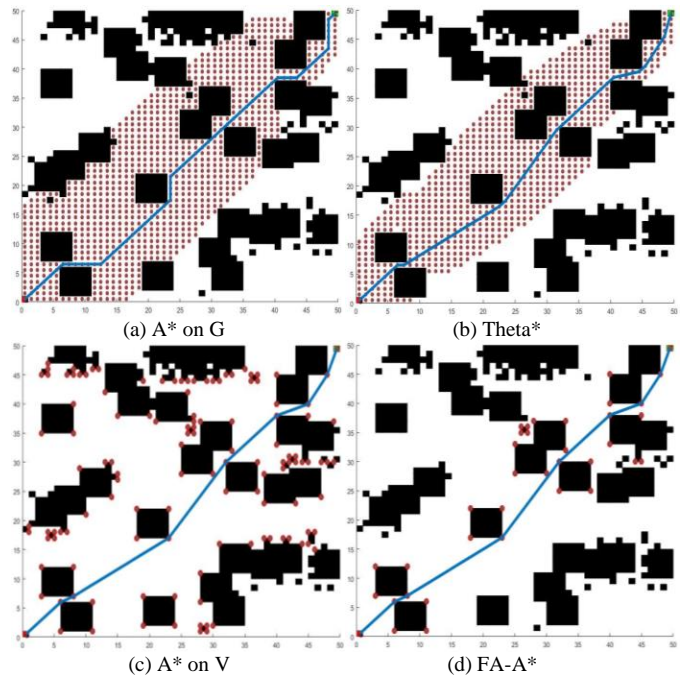

(a) A* on G    (b) Theta*

(c) A* on V    (d) FA-A*

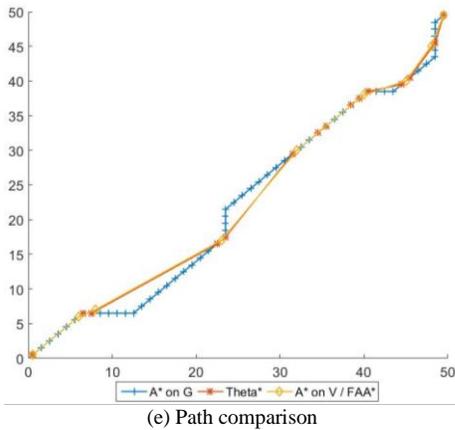

(e) Path comparison

Fig. 19. Paths found and nodes evaluated by each algorithm (300*300, 35 clusters)

## D. Maze Routing

We further validate the findings reported in the preceding sub-sections by using one of the benchmark maze routing problems [3]. It is a 511*511 map with 2.79% obstacles and 17 clusters. The starting node and the target node are chosen as (15, 466) and (466, 15). In preceding tests, the set $V_1$ is comprised of the vertices that belong to the convex hulls of clusters. In this test, we use the vertices on the convex corners instead, because clusters are nested together in such maze routing problems. The vertices on the convex hull of a cluster are essentially a simplified rendition of the vertices on the convex corners of a cluster. The term 'convex corner' means that the inner angle of the corner is less than 180 degrees. The results are shown in Table 5 and Fig. 20. FA-A* finds the shortest path among the four algorithms. Besides, as can be seen in Fig. 21, A* on Grids and Theta* are 'trapped' in the maze and evaluate a huge number of nodes throughout (149,689 and 147,896) whereas FA-A* only evaluates 104. Hence, FA-A* is 98.71% faster than A* on G and 99.23% faster than Theta*. Additionally, consistent with all previous test cases, FA-A* has the least node evaluations as shown in Fig. 22, which gains the algorithm advantage in computational time. But the runtime edge of evaluating less nodes in this test case is offset by Candidate Vertices calculations which serves as the reason that FA-A* is slower than A* on V by 0.42%. However, if the number of cluster increases, FA-A* could easily surpass A* on Visibility Graphs regarding the runtime as illustrated in the pre-ceding subsections.

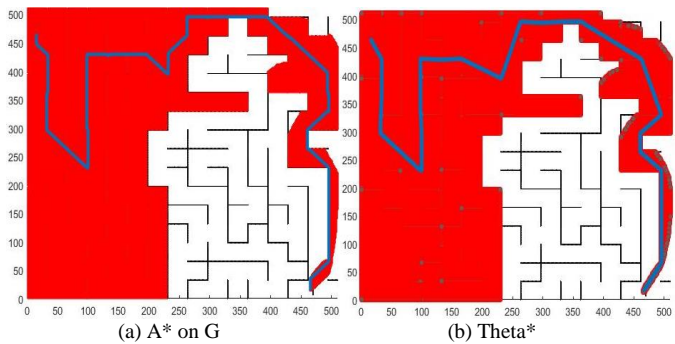

(a) A* on G  (b) Theta*

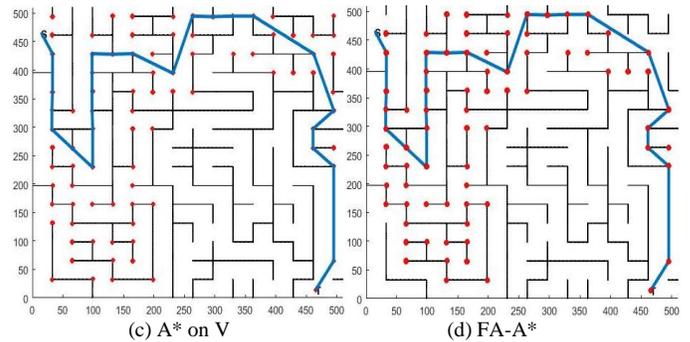

(c) A* on V  (d) FA-A*

Fig. 20. Path found for 511*511 maze routing

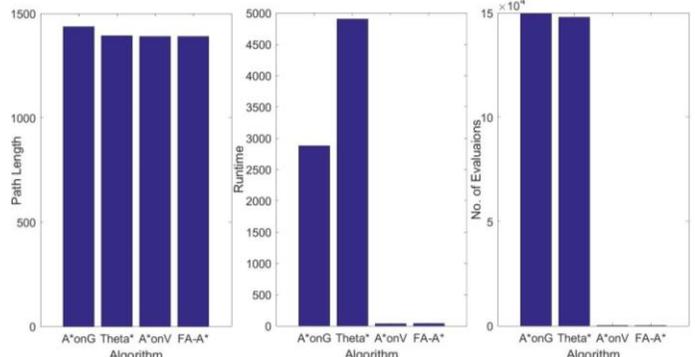

Fig. 21  511*511 maze routing

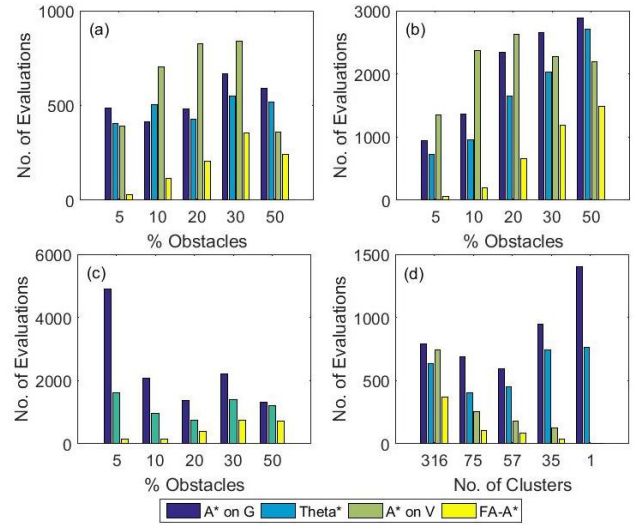

(a) 50*50  (b) 100*100
(c) 300*300  (d) 50*50 with different number of clusters

Fig. 22. Number of evaluations comparison

Table 5 Performance on a 511*511 maze

| Map Info | Algorithm | L | No. | T |
|---|---|---|---|---|
| 511*511 | A* on G | 1436.8 | 149689 | 2877.5 |
| 2.79% | Theta* | 1393.9 | 147896 | 4901.4 |
| 17 clusters | A*on V true shortest | 1389.9 | 120 | 36.9722 |
| | FA-A*(w=1) | 1389.9 | 104 | 37.1282 |

## VI. CONCLUSION

In this research, we develop a focal any-angle A* algorithm based on visibility graphs (FA-A*). FA-A* performs the path-finding with focuses on position of the target and prunes the full visibility graph accordingly. The algorithm enhances the computational performance of A* on Visibility Graphs as demonstrated in the numerical tests. Meanwhile, it always finds better paths than A* on Grids and Theta*. Moreover, FA-A* has the least node evaluations for all the test cases. While in this paper all the path costs are evaluated as Euclidean distances, for problems that have more complicated path costs, FA-A* would gain more significant advantage in terms of computational time. FA-A* not only can preserve the optimality of visibility graph-based methods but also can keep up with grid-based methods computationally. As a result, the newly developed approach is capable of finding better paths than A* and Theta*, and yields better computational efficiency than A* on Visibility Graphs or even A* and Theta* for certain cases. The algorithm can be utilized in a variety of applications, e.g., to find the optimal path in a static environment for a robot to take, tackling obstacle avoidance problems for vehicles, or to find the best connections in additively manufactured components, etc.

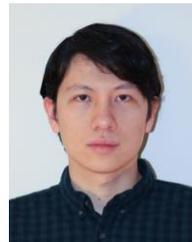
**Pei Cao** received the B.S. degree in Automation from Northwestern Polytechnical University, Xi'an, China, in 2011. He is currently pursuing the Ph.D. degree in Mechanical Engineering at University of Connecticut, Storrs, USA. He joined the Dynamics, Sensing, and Controls Laboratory at UConn in August, 2012. His research interests include global optimization, dynamic programming, statistical inference, layout design, and machine learning.

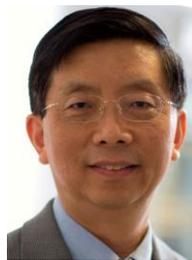
**Robert X. Gao** (M'91–SM'00–F'08) received the Ph.D. degree from the Technical University of Berlin, Berlin, Germany, in 1991. He is a Professor of Mechanical & Aerospace Engineering, and holds the Cady Staley Endowed Chair at the Case Western Reserve University, Cleveland, OH, USA. His research interests include physics-based sensing methodology, mechatronic systems design and characterization, energy harvesting, wireless sensor networks, and non-stationary signal processing. Dr. Gao is an Associate Editor for the *ASME Journal of Manufacturing Science and Engineering* and the *IFAC Journal of Mechatronics*. His research and service to the community have been recognized by multiple awards, including the IEEE Instrumentation and Measurement Society's Technical Award, NSF CAREER award, multiple best paper awards, Outstanding Associate Editor award from the IEEE Transactions on Instrumentation and Measurement, etc. He is a fellow of the SME and ASME, an Associate Member of the International Academy for Production Engineering (CIRP), a Distinguished Lecturer of the IEEE Instrumentation and Measurement Society, and a member of the Connecticut Academy of Science and Engineering.

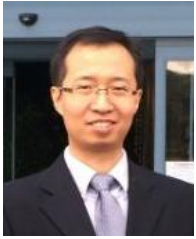

**Zhaoyan Fan (M'05)** received the B.S. degree in mechanical engineering from Tsinghua University, Beijing, China, in 2000, the M.S. degree in physical electronics from the Institute of Electronics, Chinese Academy of Sciences, Beijing, in 2003, and the Ph.D. degree in mechanical engineering from the University of Massachusetts Amherst, Amherst, MA, USA, in 2009. He joined the Mechanical, Industrial and Manufacturing Engineering Department of Oregon State University in 2016 where he is currently an Assistant Professor. His research interests include physical sensing methodology and sensing electronics, energy-efficient sensor networks, imaging methods for industrial system monitoring, smart sensors and structures, and numerical modeling/analysis for the mechatronic system.

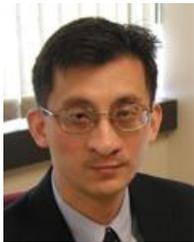

**Jiong Tang (M' 09)** received the B.S. and M.S. degrees in Applied Mechanics from Fudan University, China, in 1989 and 1992, respectively, and the Ph.D. degree in Mechanical Engineering from the Pennsylvania State University, USA, in 2001. He worked at the GE Global Research Center as Mechanical Engineer during 2001 to 2002. He then joined the Mechanical Engineering Department, University of Connecticut where he is currently a Professor and the Director of Dynamics, Sensing, and Controls Laboratory. His research interests include structural dynamics and system dynamics, control, and sensing and monitoring. Dr. Tang served as an Associate Editor for IEEE Transactions on Instrumentation and Measurement from 2009 to 2012. He also served as an Associate Editor for ASME Journal of Vibration and Acoustics, and an Associate Editor for ASME Journal of Dynamic Systems, Measurement, and Control.